\documentclass[runningheads]{llncs}

\usepackage{booktabs} % For formal tables
\usepackage{multirow}
\usepackage{algorithm}
\usepackage[noend]{algpseudocode}
\usepackage{amsmath}
\usepackage{amsfonts}
\usepackage{placeins}
\usepackage{subcaption}
\usepackage{graphicx}
\usepackage{listings}
\usepackage{xcolor}

\begin{document}

\title{LUMOS: Large User MOdel Series for Multi-Task User Behavior Prediction}

\author{Dhruv Nigam\inst{1} \and Naman Agarwal\inst{2} \and Krishna Murthy\inst{1} \and Susmit Saha\inst{1}}

\institute{Dream11, India\\
\email{dhruv.nigam@dream11.com}\\
\email{krishna.murthy@dream11.com}\\
\email{susmit.saha@dream11.com}
\and
Dream Sports, India\\
\email{naman.agarwal@dreamsports.group}}

\maketitle

\begin{abstract}
User behavior prediction at scale remains a critical challenge for online B2C platforms. Traditional approaches rely heavily on task specific models and domain specific feature engineering. This is time-consuming, computationally expensive, and requires domain expertise and therefore not scalable. We present LUMOS (Large User MOdel Series), a transformer-based architecture that eliminates task specific models and manual feature engineering by learning multiple tasks jointly using only raw user activity data. LUMOS introduces a novel cross-attention mechanism that conditions predictions on future known events (e.g., holidays, sales, etc.), enabling the model to predict complex behavior patterns like "how will upcoming holidays affect user engagement?" The architecture also employs multi-modal tokenization, combining user activities, event context, and static user demographic attributes into rich representations processed through specialized embedding pathways. 

Through extensive experiments on a production dataset spanning 1.7 trillion user activity tokens from 250 million users, we demonstrate that LUMOS achieves superior performance compared to traditional task-specific models. Across 5 tasks with established baselines, we achieve an average improvement of 0.025 in ROC-AUC for binary classification tasks and 4.6\% reduction in MAPE for regression tasks. Online A/B testing validates these improvements translate to measurable business impact with a 3.15\% increase in Daily Active Users.

\end{abstract}

% Introduction must fit on first page
\section{Introduction}

Consider a sports engagement platform preparing for the annual Indian Premier League (IPL) cricket tournament. The platform needs to predict which of its existing users will engage with the platform, the nature and extent of their engagement, and level of activity for each user during the tournament to drive personalization efforts. Traditional approaches would involve training one model per task - each requiring manual feature engineering - calculating features like "days\_since\_last\_activity," "average\_session\_length," or "activity\_during\_last\_season." This approach requires a lot of domain knowledge and is time-consuming and expensive to maintain. It also leads to duplication of effort and a lack of generalization across tasks.

We propose a shift in user behavior modelling inspired by the trends in NLP where the field moved from task-specific models and feature based approaches(like bag of words and TF-IDF) to foundation models like BERT~\cite{devlin2019bert} and GPT~\cite{radford2018improving} which are not just easier to train(in terms of pre-processing steps) but also achieve better performance when data scale is sufficient. 

\begin{figure}[htbp]
    \centering
    \includegraphics[width=0.95\columnwidth]{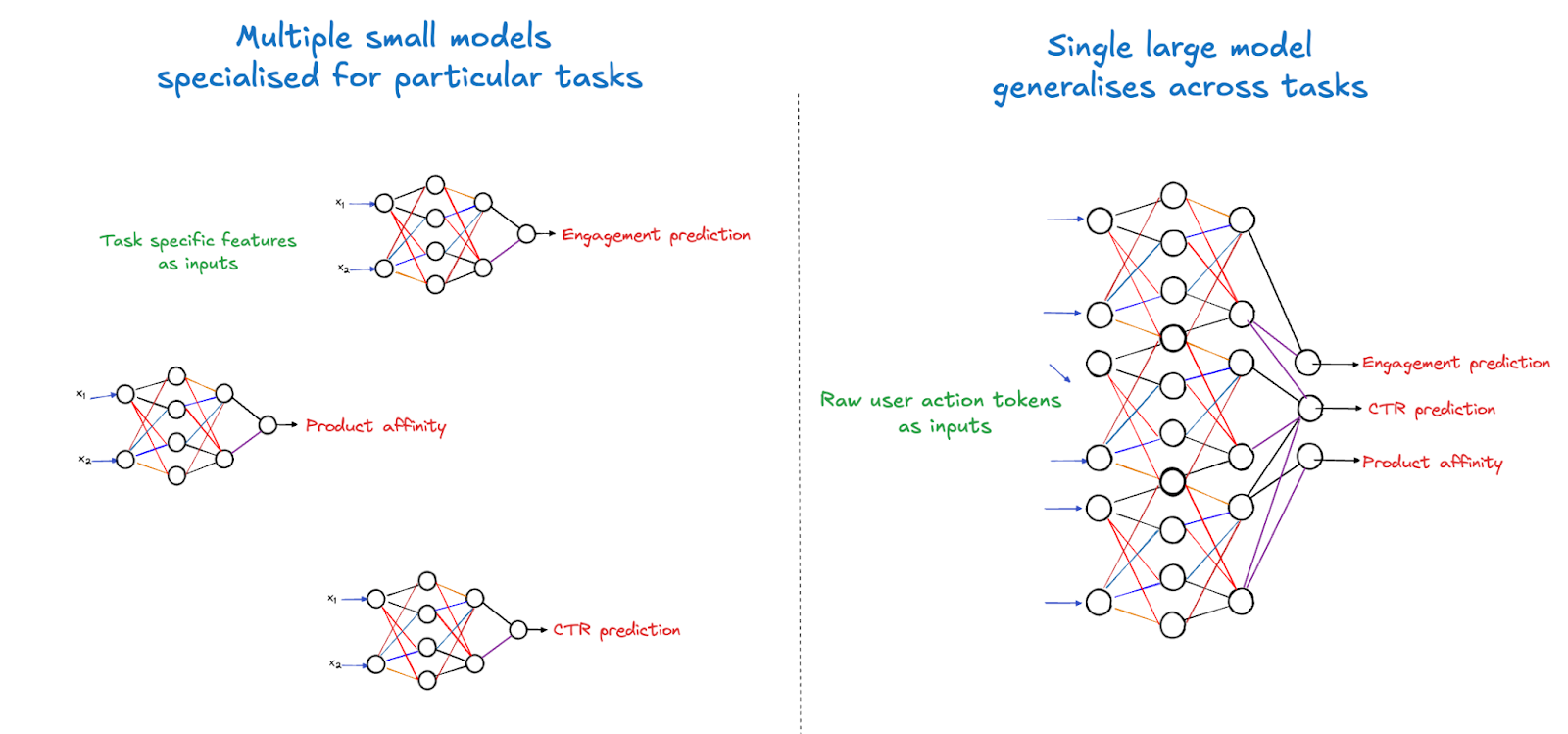}
    \caption{\small Paradigm shift from multiple specialized models to a single unified model. Traditional approaches require separate models for each prediction task (left), while LUMOS employs a single powerful model capable of handling multiple tasks simultaneously (right).}
    \label{fig:paradigm_shift}
\end{figure}

In this spirit, we present LUMOS, a transformer-based architecture that can be trained to predict future user behaviour using raw sequence of past behaviour. It eliminates manual feature engineering by learning end-to-end representations from raw user activity data. Instead of predicting a single user behaviour attribute for a pre-specified time horizon, LUMOS predicts a sequence of user behaviour tokens for multiple future time steps. Each predicted token contains the full set of predicted user behaviours precluding the need for multiple specialized models.
Experiments on our production dataset show that LUMOS achieves an average improvement of 0.025 in AUC for classification tasks and 4.6\% reduction in MAPE for regression tasks compared to the baseline task-specific models.

LUMOS uses a cross-attention mechanism between historical user behavior and future event context to answer nuanced questions like "Given that the IPL final is next Sunday, which days in the user's history are most relevant for predicting their engagement?" Our experiments show that the model correctly attends not only to recent activity but also to specific historical events like previous year's IPL finals or World Cup matches, learning these patterns entirely from data without explicit programming. This is crucial for event driven platforms like sports streaming platforms or ticketing services.

% Background and context
\section{Background}

User behavior modeling encompasses a range of tasks critical to online platforms: predicting churn, forecasting engagement levels, anticipating user actions, and understanding activity patterns~\cite{ricci2011introduction}. These predictions typically span multiple time horizons (7-day, 28-day, annual) and serve different business functions from personalized marketing to user retention strategies.

User activity exhibits characteristics like sparsity in time, sensitivity to exogenous factors like sales, festivals, events and extremely domain and task specific dynamics. 
Traditional approaches address these challenges through building separate models for each user modelling tasks. This creates a lot of redundancy since the datasets and features are often shared across tasks and the tasks themselves are often correlated. 
Additionally  data scientists may manually craft features for each task like rolling averages, recency indicators, and domain-specific aggregations. 
These factors make the process of building and maintaining these models time-consuming and expensive.

The transformer architecture, introduced by Vaswani et al.~\cite{vaswani2017attention}, has become the foundation for modern Large Language Models (LLMs) due to its ability to model long-range dependencies in sequential data. The architecture's core components include a self-attention mechanism, which allows the model to weigh the importance of different parts of the input sequence; multi-head attention, which captures diverse patterns in parallel; and positional embeddings, which preserve temporal ordering. While the original design included an encoder and a decoder for machine translation, most modern LLMs, such as GPT~\cite{radford2019language} and LLaMA~\cite{touvron2023llama}, adopt a decoder-only architecture optimized for next-token prediction.

While transformers excel at natural language processing, adapting them to user behavior prediction presents unique challenges:

\textbf{1. Continuous vs. Discrete Temporal Patterns}: Text has natural boundaries (characters, tokens or words), while user behavior is a continuous stream with irregular patterns across time. A user might be active daily, then disappear for months. We thus need to come up with a tokenization scheme that can capture the user's activity in a time window t.

\textbf{2. Multi-Modality of tokens}: Language models operate on discrete tokens from a fixed vocabulary. User behavior data consists of continuous numerical features (session lengths, activity counts) requiring different embedding strategies. Additionally user activity in a time window t is informative only in the event context of time t. For example, high activity on a days where there are matches of a specific team implies that the user is a fan of that team and is likely to be engaged only when their team plays.
This means that tokens need to be multi-modal, i.e. they need to contain information about the user's activity in the time window t as well as the event context of time t.

\textbf{4. Known Future Context}: Unlike language generation, user behavior prediction often involves known future events (scheduled matches, planned promotions) that influence predictions. Decoder only models like GPT are not able to attend to the future context since they are trained to predict the next token in a sequence given the previous tokens.

\textbf{5. Multi-Task Objectives}: Unlike text where Transformers are trained to predict the next token in a sequence - a single ,multi class classification problem, user behaviour prediction is not a single task but a multi-task problem. Typically we want to predict multiple tasks (churn, activity levels, engagement) with varying nature of predictions (binary, continuous) for different time horizons from the same user representation.

These challenges motivate the design decisions in LUMOS, particularly the novel tokenization scheme, cross-attention mechanism, and multi-task learning framework described in the next section.

% Core technical contribution
\section{Method}

\subsection{Problem Formulation}

We formulate user behaviour prediction as a sequence modelling problem. We chose a day as the unit of time since most tasks are defined at a daily level. Given a user's historical data up to day $t$, we aim to predict multiple behavioural outcomes over $T_{fut}$ future days.

Inputs to the model are:
\begin{itemize}
    \item $\mathbf{U}^{hist} \in \mathbb{R}^{T_{hist} \times d_u}$: Historical user activity features over $T_{hist}$ days. For each day $t$, $\mathbf{u}_t \in \mathbb{R}^{d_u}$ contains the user's aggregated activity features for that day.
    \item $\mathbf{S}^{hist} \in \mathbb{R}^{T_{hist} \times d_s}$: Historical event context (e.g., match schedules, sporting calendar). For each day $t$, $\mathbf{s}_t \in \mathbb{R}^{d_s}$ contains the raw event context vector for that day.
    \item $\mathbf{S}^{fut} \in \mathbb{R}^{T_{fut} \times d_s}$: Future event context (known upcoming events)
    \item $\mathbf{x}^{static} \in \mathbb{R}^{d_{static}}$: Static user features (demographics)
\end{itemize}

The model outputs future behavior tokens $\hat{\mathbf{U}}^{fut} \in \mathbb{R}^{T_{fut} \times d_u}$ for each of the next $T_{fut}$ days. For each future day $t$, $\hat{\mathbf{u}}_t \in \mathbb{R}^{d_u}$ contains the predicted user's aggregated activities for that day. Each token contains the full set of predicted user behaviours, mirroring the dimensionality and structure of the input user features $\mathbf{U}^{hist}$.

\subsection{Sequence Construction and Temporal Windows}

We construct training sequences using fixed temporal windows of $T_{hist} = 360$ days for historical context and $T_{fut} = 7$ days for future predictions. The 360-day window captures the cyclical nature of the sports calendar, allowing the model to learn annual patterns such as recurring tournaments and league seasons, while the 7-day horizon aligns with practical short-term prediction needs in applications like churn prevention and product recommendations. User activity is inherently sparse, so for days without activity we set $\mathbf{u}_t = \mathbf{0}$ while still providing event context $\mathbf{s}_t$ and static features $\mathbf{x}^{static}$, enabling the model to learn disengagement patterns as a function of available events. For users with fewer than 360 days on the platform, we pad the sequence with a learned padding token $\mathbf{u}_t = \mathbf{u}_{pad}$ (distinct from $\mathbf{0}$) for days before registration, allowing the model to distinguish between pre-registration periods and days with zero activity.

\subsection{LUMOS Architecture Overview}

\begin{figure}[htbp]
    \centering
    \includegraphics[width=1.1\columnwidth]{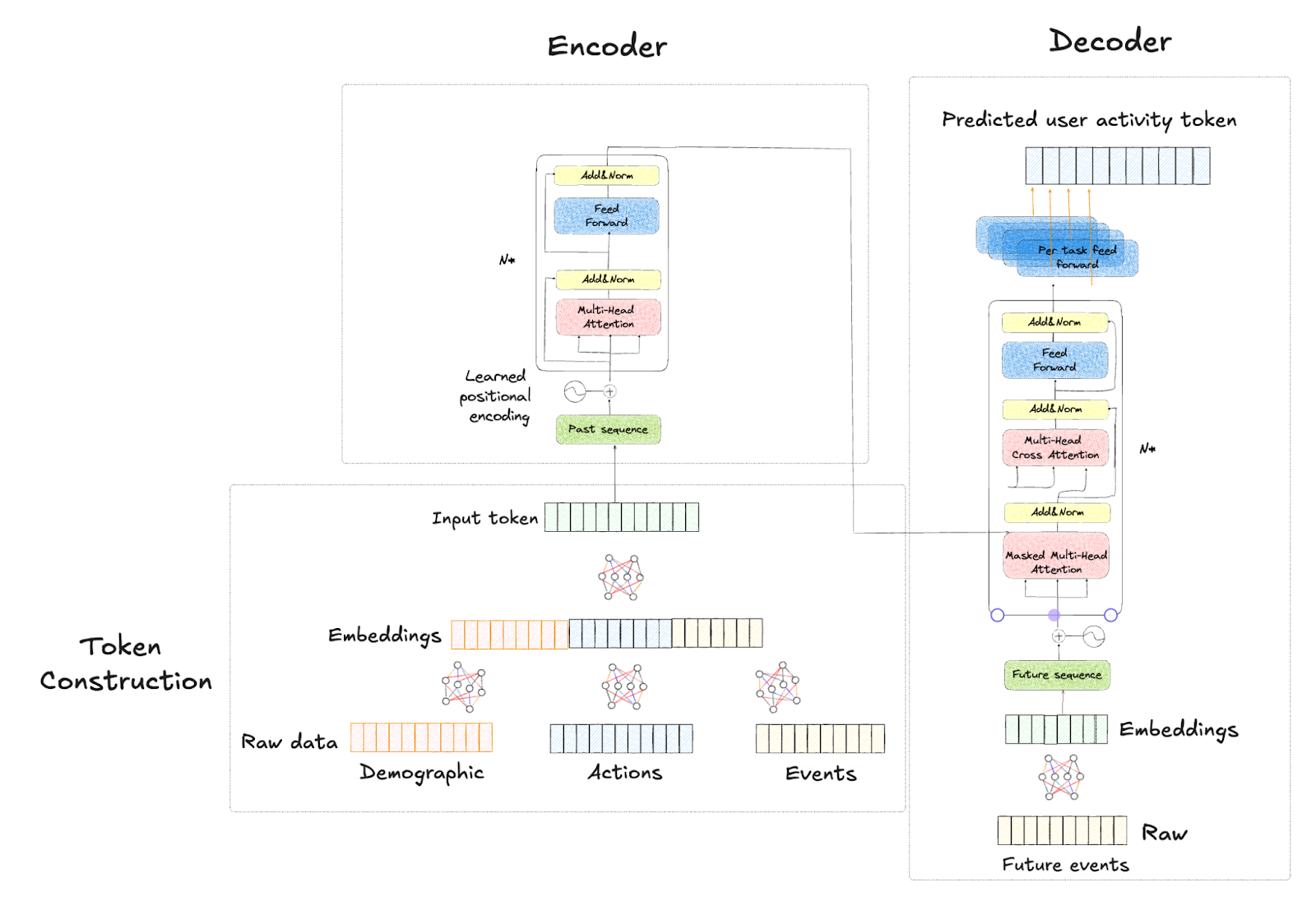}
    \caption{\small LUMOS architecture showing the complete model pipeline from token construction to prediction. The token construction process (bottom) combines user activities, event context, and static features into unified daily tokens. The encoder processes the historical token sequence through self-attention, producing contextualized representations of the same dimension as the input tokens. The decoder uses cross-attention where future event context tokens query the encoded historical representations to generate future-conditioned embeddings. Finally, an output projection layer maps each decoder output to predicted user behavior dimensions $\mathbf{U}^{fut}$.}
    \label{fig:architecture}
\end{figure}

LUMOS employs an encoder-decoder transformer architecture with specialized components for user behavior modeling. Figure~\ref{fig:architecture} illustrates the complete architecture. We now detail each component.

\subsubsection{Multi-Modal Token Construction}

To address the challenges of tokenizing continuous user behavior (Section 2), we adopt a novel multi-modal tokenization scheme. We aggregate user activity $\mathbf{U}^{hist}$ and event context $\mathbf{S}^{hist}$ at a daily level, yielding discrete, equally-spaced temporal tokens. Since we have three modalities—user activity, event context, and static features—we embed each modality separately using configurable MLPs, then fuse them via a cross-modal interaction layer.Each modality is processed by a dedicated MLP with configurable depth (1-3 layers, we use depth=2). These embeddings transform raw numerical features into learned representations:

\begin{align}
\mathbf{h}^{static} &= \text{MLP}_{static}(\mathbf{x}^{static}) \in \mathbb{R}^{d_{static\_embed}} \\
\mathbf{h}_t^{user} &= \text{MLP}_{user}(\mathbf{u}_t) \in \mathbb{R}^{d_{user\_embed}} \\
\mathbf{h}_t^{event} &= \text{MLP}_{event}(\mathbf{s}_t) \in \mathbb{R}^{d_{supply\_embed}}
\end{align}

\textbf{Cross-Modal Interaction Layer.} Unlike simple concatenation, we pass the concatenated embeddings through a final projection MLP that learns interactions between modalities. For example, a user's high activity level during IPL finals has different meaning than the same activity during off-season. This layer captures such contextual dependencies:

\begin{equation}
\mathbf{z}_t = \text{MLP}_{proj}([\mathbf{h}_t^{user}; \mathbf{h}^{static}; \mathbf{h}_t^{event}]) \in \mathbb{R}^{d_{model}}
\end{equation}

where $[\cdot ; \cdot]$ denotes concatenation and static features are broadcast across all timesteps.

\textbf{Future Token Construction.} For future timesteps, we only have event context: $\mathbf{z}_t^{fut} = \text{MLP}_{event}^{fut}(\mathbf{s}_t^{fut})$, using a separate embedding layer to learn distinct representations for future versus historical event context. The token construction pipeline is illustrated in Figure~\ref{fig:architecture}.

\subsubsection{Temporal Encoding}

Temporal position is critical for user behavior modeling, as recent activity typically matters more than distant history, though specific past events (e.g., last year's tournament finals) can be highly relevant. While language models conventionally rely on sinusoidal embeddings~\cite{vaswani2017attention} or rotation-based encodings like RoPE~\cite{su2021roformer}, position encoding behavior is highly domain-specific.

Through systematic comparison (Table~\ref{tab:positional} in Appendix~\ref{sec:positional_analysis}), we found learned position embeddings yield the best performance. The embeddings are initialized to zero, added to token sequences, and optimized during training. Fourier analysis of the learned embeddings reveals they primarily encode low-frequency temporal patterns (see Appendix~\ref{sec:positional_analysis} for detailed frequency analysis). This suggests that for user behavior, broad temporal context (recent week vs. distant past) matters more than precise day-level positioning, contrasting with language models where exact token position is critical for syntax and semantics.

\subsubsection{Encoder: Learning from History}

The encoder stack processes the sequence of historical tokens $\mathbf{H}^{enc} = \text{Encoder}(\mathbf{Z}^{hist} + \mathbf{PE}^{hist})$ following standard transformer architecture. We employ pre-normalization~\cite{xiong2020layernormalizationtransformerarchitecture} for training stability and SwiGLU activation functions~\cite{shazeer2020glu}, which have demonstrated superior performance compared to traditional ReLU or GELU activations. 

\subsubsection{Decoder: Conditioning on the Future}

The decoder is where LUMOS diverges most significantly from standard LLMs. Instead of generating predictions for the next token using the encoder representation of the final input token, our decoder processes future event context tokens to condition predictions. This is achieved through a cross-attention mechanism between the future event context tokens and the encoder output sequence. Critically, we remove self-attention from the decoder layers, ensuring that each future time step derives its final representation solely from the historical context (via cross-attention to the encoder) and its own event context.

Mathematically, the decoder processes future event context embeddings (augmented with positional encodings) through cross-attention to the encoder outputs:

\begin{equation}
\mathbf{H}^{dec} = \text{Decoder}(\mathbf{Z}^{fut} + \mathbf{PE}^{fut}, \mathbf{H}^{enc})
\end{equation}

where $\mathbf{PE}^{fut} \in \mathbb{R}^{T_{fut} \times d_{model}}$ are learned positional embeddings for the future time steps. For each decoder layer, the cross-attention mechanism computes:

\begin{align}
\mathbf{Q} &= (\mathbf{Z}^{fut} + \mathbf{PE}^{fut}) \mathbf{W}_Q \quad \text{(queries from future event context)} \\
\mathbf{K} &= \mathbf{H}^{enc} \mathbf{W}_K \quad \text{(keys from encoded history)} \\
\mathbf{V} &= \mathbf{H}^{enc} \mathbf{W}_V \quad \text{(values from encoded history)} \\
\text{CrossAttn}(\mathbf{Q}, \mathbf{K}, \mathbf{V}) &= \text{softmax}\left(\frac{\mathbf{Q}\mathbf{K}^T}{\sqrt{d_k}}\right) \mathbf{V}
\end{align}

where $\mathbf{W}_Q, \mathbf{W}_K, \mathbf{W}_V$ are learned projection matrices. The key distinction from standard transformer decoders is that queries come from the future event context embeddings while keys and values come from the historical encoder representation.

\subsubsection{User Behaviour Token Prediction}
The final step transforms decoder outputs into predicted user behaviour tokens. For each future time step $t$, the decoder produces a representation $\mathbf{h}_t^{dec} \in \mathbb{R}^{d_{model}}$, which is mapped to predictions through task-specific projection heads. These projection heads consist of separate feed-forward networks for each of the $d_u$ behavioural dimensions. Each projection head comprises two hidden layers of dimension $d_{model}$ with ReLU activation, followed by a single output layer that produces a scalar prediction. The outputs from all $d_u$ projection heads are concatenated to form the predicted user behaviour tokens $\hat{\mathbf{U}}^{fut} \in \mathbb{R}^{T_{fut} \times d_u}$.

Each predicted token $\hat{\mathbf{u}}_t \in \mathbb{R}^{d_u}$ maintains identical dimensionality to the historical user tokens, containing the same aggregated activity features across all behavioural dimensions.

\subsection{Training Objective}

We optimize a loss function that compares predicted user behaviour tokens against the ground truth user behaviour tokens. We employ uncertainty-based weighting \cite{kendall2018multi} to automatically balance the contribution of each user behaviour dimension. Specifically, for a multi-task loss with $d_u$ user behaviour dimensions, we formulate the weighted loss as:

\begin{equation}
\mathcal{L} = \sum_{i=1}^{d_u} \frac{1}{2\sigma_i^2}\mathcal{L}_i(\mathbf{U}^{fut}_i, \hat{\mathbf{U}}^{fut}_i) + \log\sigma_i^2
\end{equation}

where $\mathcal{L}_i$ is the task-specific loss for the $i$-th behavior dimension. The loss function uses binary cross-entropy (BCE) for binary behaviors such as churn and reactivation, and mean squared error (MSE) with zero-masking for continuous behaviors such as average session length and other continuous engagement metrics. The zero-masking ensures that days with no activity do not contribute to the loss for continuous metrics.

The parameter $\sigma_i^2$ is a learnable parameter representing the uncertainty associated with behavior dimension $i$. The term $\frac{1}{2\sigma_i^2}$ serves as a dynamic weight that inversely scales with task uncertainty, while the regularization term $\log\sigma_i^2$ prevents the model from assigning zero uncertainty to any dimension. During training, the model jointly optimizes both the prediction accuracy and these uncertainty parameters, allowing it to dynamically adjust focus across different aspects of user behavior based on their relative difficulty and importance.

\subsection{Model Specifications}
\label{subsec:model_card}

Table~\ref{tab:model_card} provides complete specifications for the production LUMOS model.
\begin{table}[h]
\centering
\caption{LUMOS Model Card}
\label{tab:model_card}
\small
\begin{tabular}{ll}
\toprule
\multicolumn{2}{c}{\textbf{Architecture}} \\
\midrule
Type & Encoder-Decoder Transformer with Cross-Attention \\
Total Parameters & 7.23M \\
Encoder Layers & 6 layers, 8 heads, $d_{model}=512$, $d_{ff}=2048$ \\
Decoder Layers & 6 layers, 8 heads (cross-attention only) \\
Positional Encoding & Learned embeddings \\
Activation & SwiGLU (transformer), ReLU (projection heads) \\
Normalization & Pre-LayerNorm \\
\midrule
\multicolumn{2}{c}{\textbf{Input/Output Specifications}} \\
\midrule
Historical Window & $T_{hist} = 360$ days \\
Future Window & $T_{fut} = 7$ days \\
User Features & $d_u = 13$ dimensions \\
Event Context Features & $d_s = 32$ dimensions \\
Static Features & $d_{static} = 9$ dimensions \\
Token Embedding Dims & user=128, event context=64, static=32 \\
Output Predictions & $(T_{fut}, d_u)$ - 3 binary, 10 continuous \\
\midrule
\multicolumn{2}{c}{\textbf{Training Configuration}} \\
\midrule
Optimizer & AdamW (lr=$1 \times 10^{-4}$, weight decay=0.01) \\
Batch Size & 512 per GPU $\times$ 8 GPUs = 4096 effective \\
Precision & BF16 mixed precision \\
Hardware & 8$\times$ A100 40GB (p4d.24xlarge) \\
Training Time & 72 hours (576 A100 GPU-hours) \\
Dataset & 250M users, $\sim$48TB processed \\
\midrule
\bottomrule
\end{tabular}
\end{table}

\textbf{Auxiliary Outputs.} Beyond behavioral predictions, LUMOS generates reusable embeddings: (1) dynamic user embeddings ($d_{model}=512$) from encoder outputs, (2) static user embeddings ($d_{static\_embed}=32$) for cold-start scenarios, and (3) event context embeddings ($d_{supply\_embed}=64$) representing daily event context. See Appendix~\ref{sec:embeddings} for extraction methods and evaluation.

\subsection{Training Pipeline and Infrastructure}
\label{subsec:training}

\begin{figure}[htbp]
    \centering
    \includegraphics[width=\columnwidth]{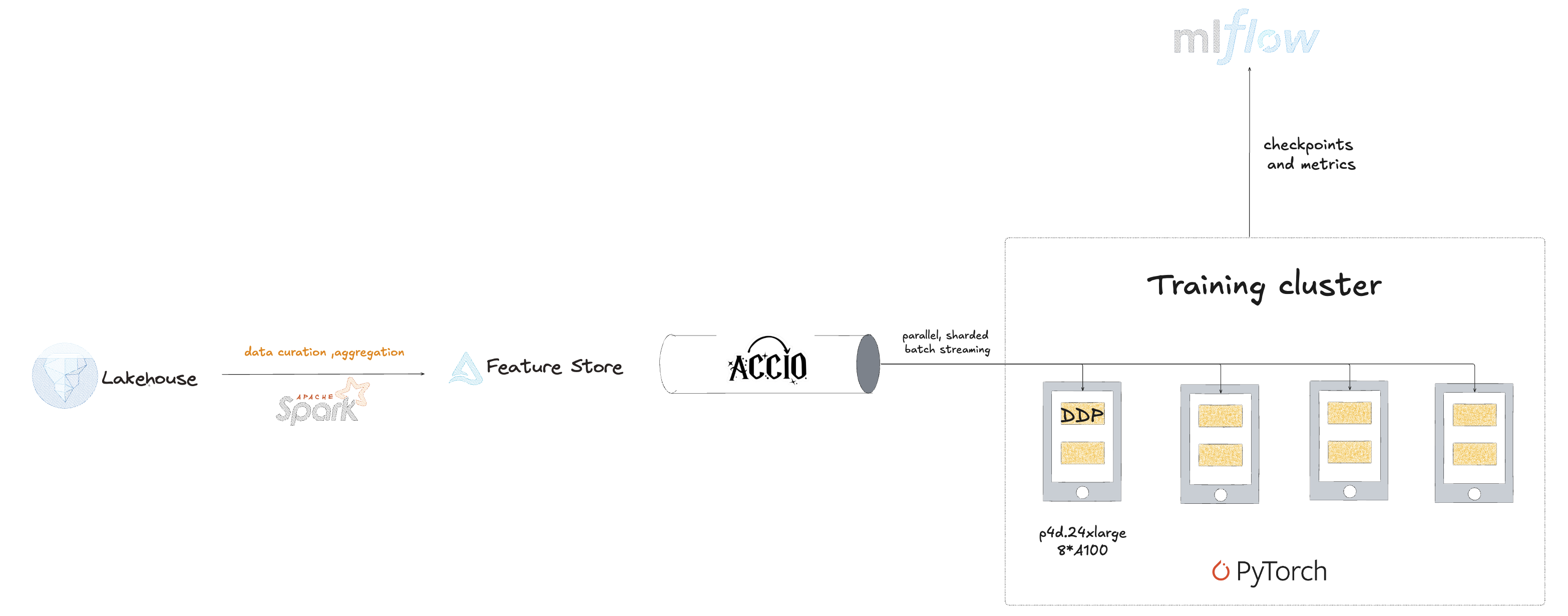}
    \caption{LUMOS training pipeline showing the data flow from raw user activities through PySpark preprocessing, Delta Lake storage, streaming data loading via Accio, and distributed training across multiple GPUs.}
    \label{fig:training_pipeline}
\end{figure}

Training LUMOS at scale requires a sophisticated data pipeline and distributed training infrastructure to handle the massive volume of user behavioral data. Figure~\ref{fig:training_pipeline} illustrates the complete training pipeline from raw data to model deployment.

\textbf{Data Scale and Preparation.} Our training dataset comprises 250 million users over extended temporal windows, resulting in ~1.7 trillion user activity tokens. Each training observation constitutes a historical sequence of up to 360 daily tokens, with each token containing aggregated activity features across multiple dimensions ($d_u = 13$ behavioral dimensions). The complete dataset spans ~50 terabytes when including historical user activities, event context sequences, and static features. To handle this scale, we employ Apache Spark for distributed data preprocessing. The resulting feature-engineered dataset is stored in Delta Lake tables, which provide ACID transaction guarantees, schema evolution, and efficient time-travel capabilities for reproducible experiments.

\textbf{Streaming Data Loader.} We developed Accio, a custom PyTorch-compatible data loader designed specifically for training large sequence models on massive datasets stored in Delta tables. Accio addresses two critical challenges: (1) the dataset is too large to fit in memory, and (2) distributed data-parallel training requires deterministic and balanced data sharding across multiple workers. Accio operates by streaming data partition-by-partition from Delta tables, with each partition representing a subset of users. It also implements worker-aware sharding, ensuring that each DDP process receives a disjoint subset of partitions without redundant data loading.

\textbf{Distributed Training Infrastructure.} We train LUMOS using PyTorch's native Distributed Data Parallel (DDP) framework on a single multi-GPU machine. Our training infrastructure consists of AWS p4d.24xlarge instances equipped with 8 NVIDIA A100 (40GB) GPUs interconnected via NVLink, providing high-bandwidth GPU-to-GPU communication essential for efficient gradient synchronization. Each training run is distributed across all 8 GPUs, with each GPU maintaining a replica of the full model and processing a disjoint shard of each batch. 

\textbf{Training Duration and Compute.} A complete training run of the full LUMOS model (with 6 encoder layers, 6 decoder layers, $d_{model} = 512$, and 8 attention heads) requires approximately 72 hours (3 days) on our 8-GPU setup. This translates to 576 A100 GPU-hours per training run ($8 \text{ GPUs} \times 72 \text{ hours}$). We employ mixed-precision training using BF16 (Brain Floating Point)~\cite{kalamkar2019study} to accelerate computation and reduce memory footprint while maintaining numerical stability.

% Experimental setup
\section{Experiments}

\subsection{Experimental Design}

We designed our evaluation to assess LUMOS's ability to predict multiple user behavior attributes simultaneously using a single unified model. THe model we trained using LUMOS predicts 13 key user behavior attributes spanning engagement patterns, activity levels, and platform interactions on a daily aggregated basis. For 5 of these 13 attributes, we had established baseline models that enabled quantitative comparison: specialized Random Forest ensembles and deep neural networks with extensive hand-crafted features. We evaluated both offline model performance metrics (3 binary classification tasks using ROC-AUC, 2 continuous prediction tasks using MAPE) and online business impact through an A/B test.

The offline evaluation uses a large-scale dataset from Dream11 comprising 1.7 trillion user activity tokens from 250 million users over four years (2021-2024). The data consists of user activity sequences $\mathbf{U}^{hist} \in \mathbb{R}^{T_{hist} \times d_u}$, external event context $\mathbf{S}^{hist} \in \mathbb{R}^{T_{hist} \times d_s}$ (sports calendar), and static user attributes $\mathbf{x}^{static} \in \mathbb{R}^{d_{static}}$. Due to the large scale, we used only 1\% of the data as holdout set for offline evaluation.

To validate whether offline performance improvements translate to measurable business impact, we deployed LUMOS predictions to feed into \textbf{Pacman}—our proprietary intelligent offers engine designed to enhance user engagement through personalized offers. Previously, \textbf{Pacman} relied on a combination of legacy specialized baseline models. We designed a controlled A/B test comparing Pacman performance using LUMOS predictions against using the legacy specialized models. For the experiment, we created two random, mutually exclusive cohorts of approximately 700,000 users each: (1) Target group receiving offers from Pacman based on LUMOS predictions, and (2) Control group receiving offers based on legacy specialized model predictions. The experiment ran for 18 days. The primary metrics measured were Daily Active Users (DAU) and cost efficiency of the offers strategy.

% Results and empirical findings
\section{Results and Analysis}

In the offline evaluation, for 5 of the 13 attributes predicted by the LUMOS model where we had established baselines for comparison — 3 binary classification tasks and 2 continuous prediction tasks — LUMOS demonstrated substantial improvements over specialized task-specific models. Across binary classification tasks, LUMOS achieved an average improvement of +0.025 in ROC-AUC, while continuous prediction tasks showed a 4.6\% reduction in MAPE. 

To validate whether offline performance improvements translate to measurable business impact, we deployed LUMOS predictions in production via \textbf{Pacman}—our proprietary intelligent offers strategy. A controlled 18-day A/B test with approximately 700,000 users per cohort demonstrated that LUMOS-powered offers led to a 3.15\% increase in Daily Active Users (DAU) while simultaneously reducing offer expenditure by 2.47\%, confirming that offline gains translate directly to enhanced user engagement and operational efficiency.

\subsection{Attention Patterns}

The cross-attention mechanism in LUMOS allows the model to attend to relevant historical event context when predicting user behaviour for upcoming events. To verify this, we visualize the attention weights from the decoder's cross-attention layer when predicting behaviour at the start of the IPL season—the most important tournament in the Indian sports calendar, spanning two months from late March to May with daily matches. For each future timestep, the model computes 360 attention weights corresponding to the 360 days in the historical context, with the relative magnitude indicating the importance of each historical day for the current prediction.

\begin{figure}[htbp]
    \centering
    \includegraphics[width=0.95\columnwidth]{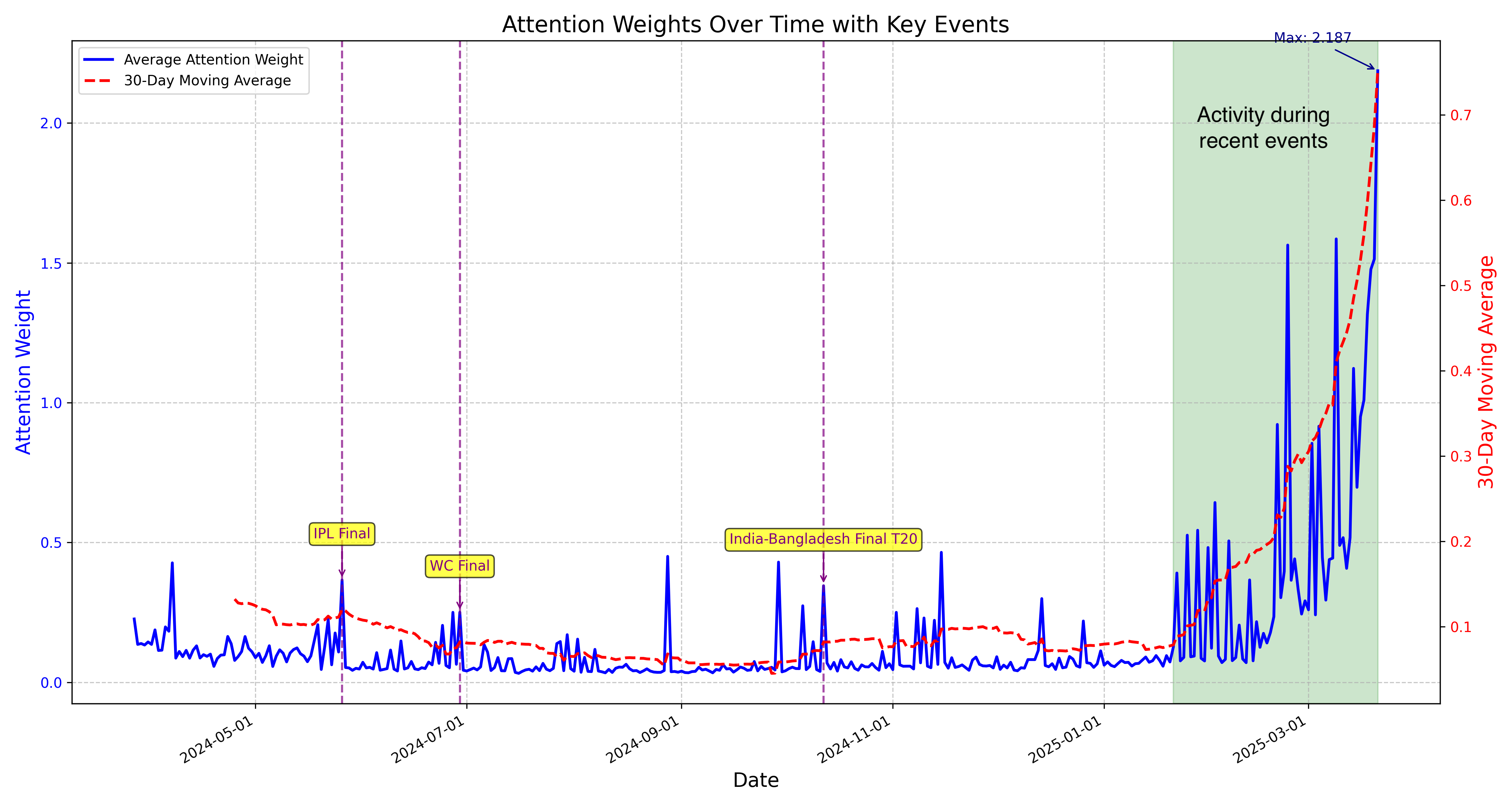}
    \caption{\small Cross-attention weights when predicting user behaviour at the start of IPL season. The heatmap reveals three distinct learned patterns: (1) \textbf{Recency focus} with strong attention to the last 7-14 days (positions 346-360), (2) \textbf{Event-specific memory} showing peaks at positions corresponding to previous year's IPL final (position 42), T20 World Cup final (position 156), and start of last IPL season (position 298), and (3) \textbf{Seasonal patterns} with moderate attention to the same calendar period from the previous year. This demonstrates that the model learns meaningful temporal dependencies without explicit programming, automatically identifying that past tournament behaviour predicts future tournament engagement.}
    \label{fig:attention}
\end{figure}

Figure~\ref{fig:attention} reveals how the model automatically learns to attend to relevant historical patterns.

\subsection{Scaling Laws}

To understand LUMOS's scaling behavior and validate its architecture against established neural network principles, we analyze how model performance scales with both training data amount and model size as was pioneered by the Chinchilla paper~\cite{hoffmann2022training}. Using a fixed 7.23M parameter model across 9 well-trained runs with data ranging from 2,532 GB to 50,000 GB, we observe a power law relationship $L(D) = 0.507 \times D^{-0.048}$ with scaling exponent $\alpha = -0.048$ ($R^2 = 0.869$, $p < 0.001$), indicating diminishing returns with increasing data—doubling data yields approximately 3.3\% loss reduction.

For model capacity, examining 4 model sizes from 7.23M to 20.81M parameters reveals a stronger scaling relationship $L(N) = 0.438 \times N^{-0.152}$ with exponent $\alpha = -0.152$ ($R^2 = 0.622$). The model size scaling exponent is approximately 3× stronger than the data scaling exponent, demonstrating that architectural capacity has greater impact on performance than data quantity in the observed regime.

\begin{figure}[htbp]
\centering
\includegraphics[width=0.45\textwidth]{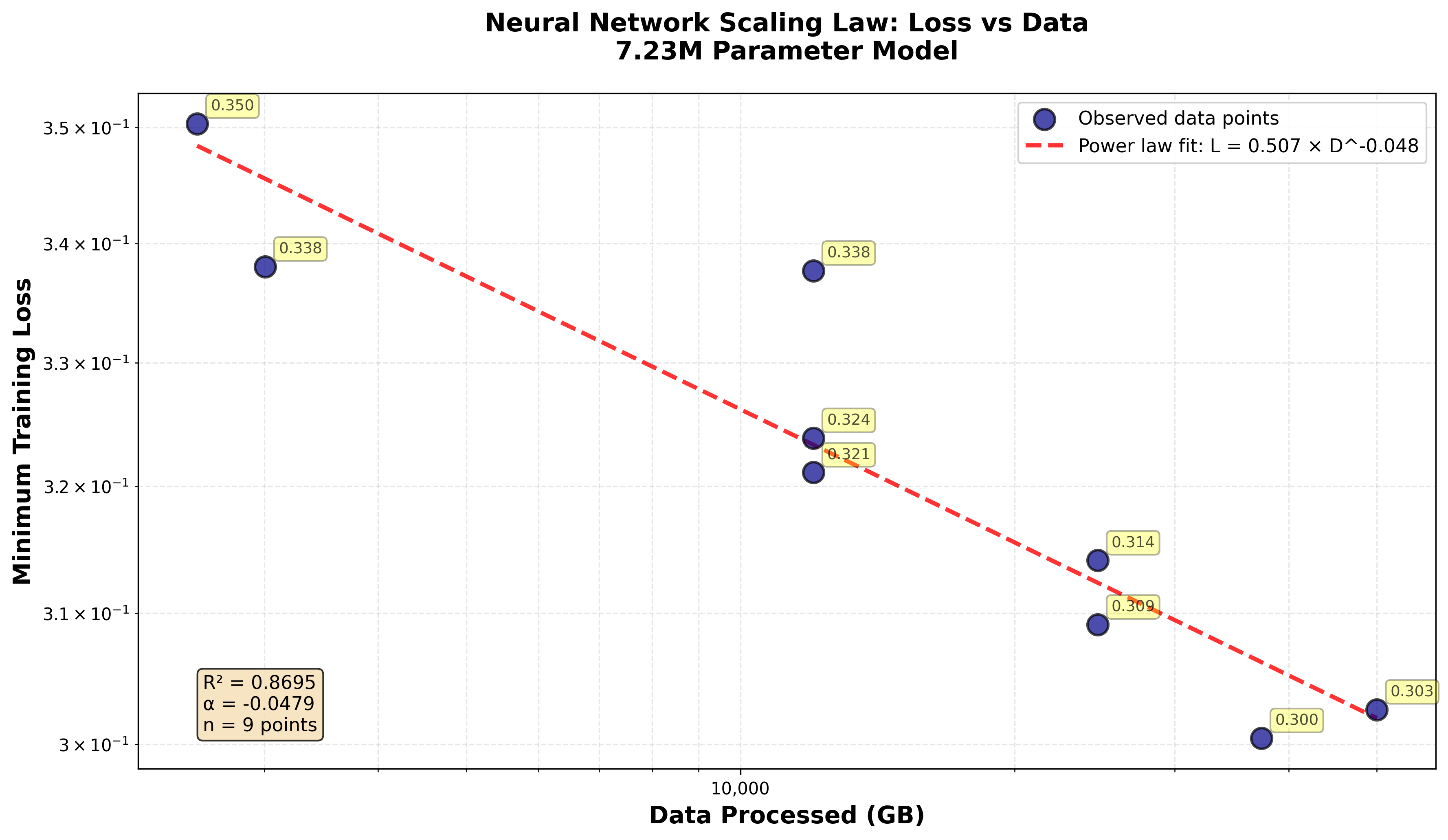}
\includegraphics[width=0.45\textwidth]{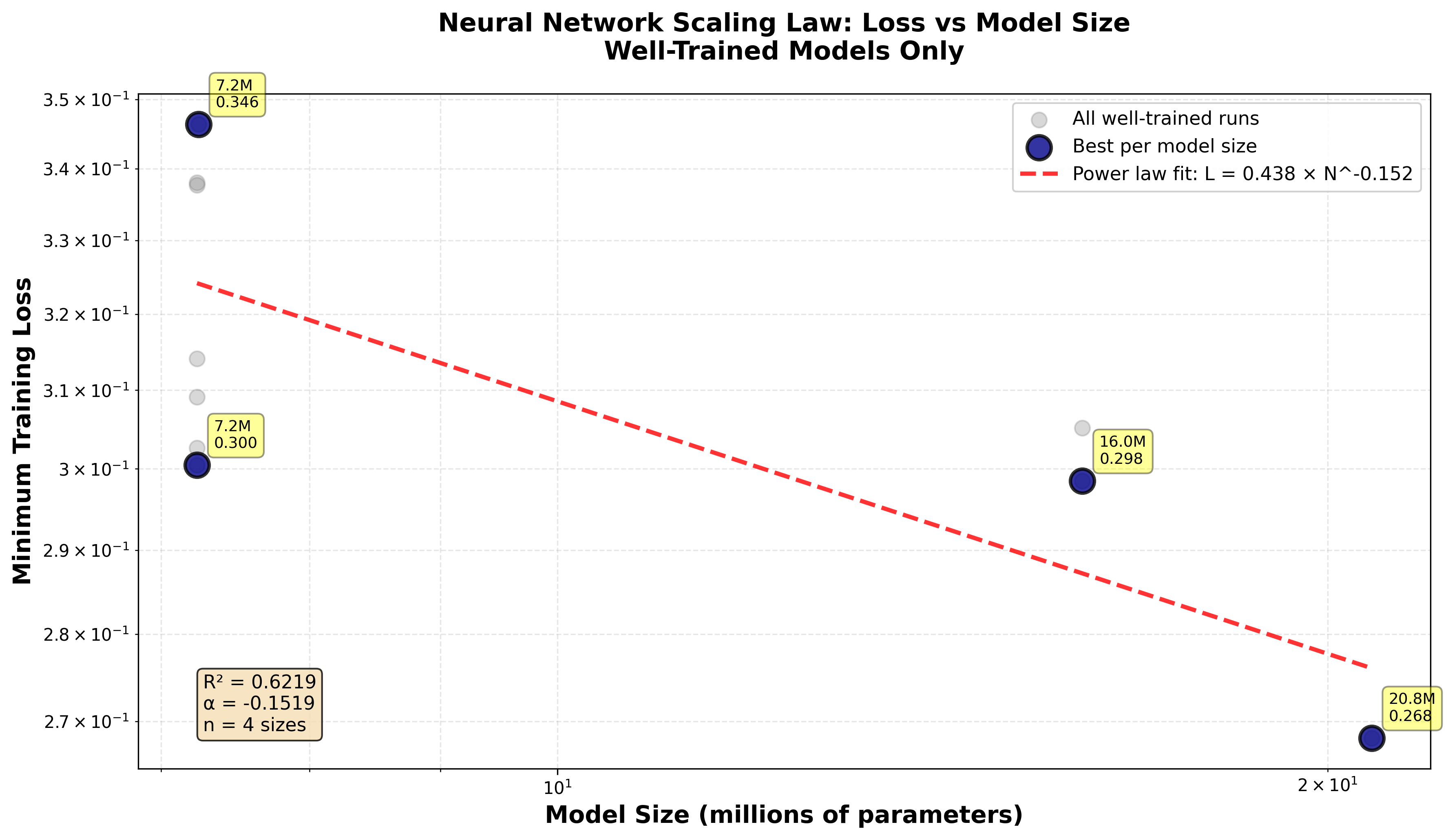}
\caption{Neural network scaling laws for LUMOS: (a) Loss vs. training data amount, (b) Loss vs. model size. Both exhibit clear power law relationships.}
\label{fig:scaling_laws}
\end{figure}

\begin{table}[h]
\centering
\caption{Model configurations tested for scaling analysis}
\label{tab:model_configs}
\begin{tabular}{lccccc}
\toprule
Config & Parameters & $d_{model}$ & Layers (E/D) & Heads & Val Loss \\
\midrule
Small & 7.23M & 512 & 6/6 & 8 & 0.3347 \\
Medium & 12.18M & 640 & 8/8 & 8 & 0.3298 \\
Large & 16.42M & 768 & 8/8 & 12 & 0.3271 \\
X-Large & 20.81M & 768 & 10/10 & 12 & 0.3254 \\
\bottomrule
\end{tabular}
\end{table}

The Chinchilla scaling laws~\cite{hoffmann2022training} found exponents of $\alpha \approx -0.34$ for data and $\alpha \approx -0.28$ for model size in language models. LUMOS exhibits weaker scaling ($\alpha_{data} = -0.048$, $\alpha_{model} = -0.152$), suggesting user behavior prediction may be more sample-efficient than language modeling but still benefits from architectural capacity.

% Conclusion and future work
\section{Conclusion}

This paper presented LUMOS, a transformer-based architecture that fundamentally rethinks user behavior prediction by eliminating manual feature engineering and specialized task-specific models in favor of end-to-end learning from raw data and unified foundation models. Through extensive offline and online experiments on millions of users, we demonstrated that LUMOS achieves state-of-the-art performance while requiring orders of magnitude less development effort for new prediction tasks.

The key technical contributions of LUMOS include: (1) a novel cross-attention mechanism that conditions predictions on future known events, enabling the model to learn complex temporal relationships between user history and upcoming event context; (2) a multi-modal tokenization scheme that unifies user activities, event context, and static features into rich daily representations; and (3) a production-ready implementation at scale - all of which are backed by empirical evidence - both offline and online.

Our results reveal several important auxiliary insights. First, the bidirectional importance of event context — where both past and future event context play a crucial role in predicting user behaviour. Second, learned positional embeddings significantly outperform generic alternatives, indicating that user behaviour exhibits unique temporal patterns requiring task-specific modeling. Third, our scaling laws demonstrate that LUMOS follows predictable power law relationships, with model size scaling ($\alpha = -0.152$) providing 3× stronger improvements than data scaling ($\alpha = -0.048$), validating the architectural approach and providing clear guidance for future scaling decisions.

As user behaviour data continues to grow in volume and complexity, approaches like LUMOS that can automatically discover patterns from raw data will become increasingly vital. By demonstrating that transformers can successfully model user behaviour at scale, we hope to inspire further research at the intersection of deep learning and user analytics. The shift from manual feature engineering to learned representations is not just a technical improvement—it fundamentally changes how we build and deploy user understanding systems, enabling more accurate, adaptable, and maintainable solutions for the challenges of modern digital platforms.

% Bibliography - splncs04 is correct for LNCS
\bibliographystyle{splncs04}
\bibliography{bibliography}

% Appendices
\appendix

% Core appendices
\section{Positional Embedding Analysis}
\label{sec:positional_analysis}

\subsection{Frequency Components Analysis}

To understand what temporal patterns the learned positional embeddings capture, we apply Fast Fourier Transform (FFT) to extract frequency components from both learned and sinusoidal embeddings.

\begin{figure}[htbp]
\centering
\includegraphics[width=0.45\textwidth]{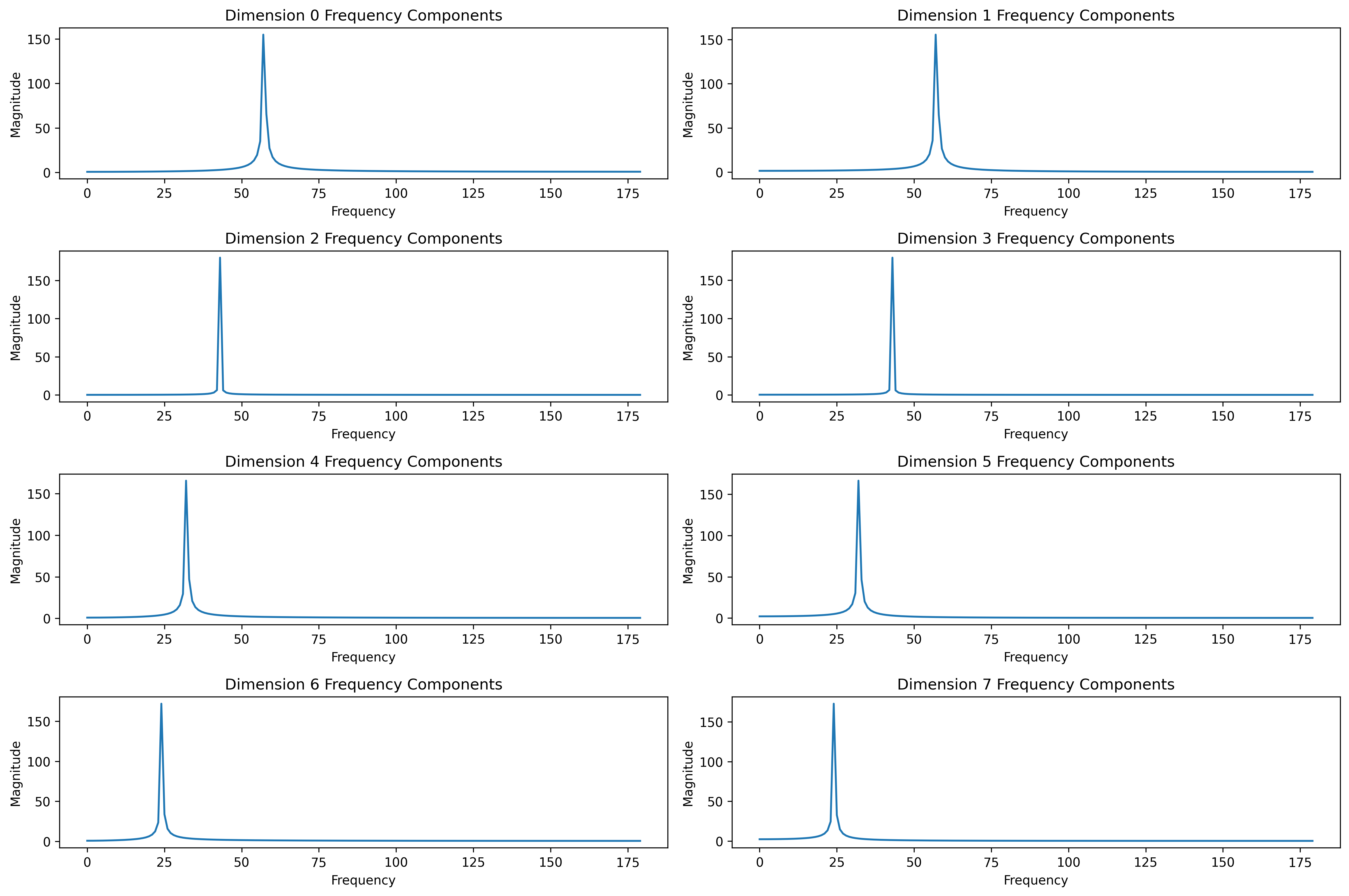}
\includegraphics[width=0.45\textwidth]{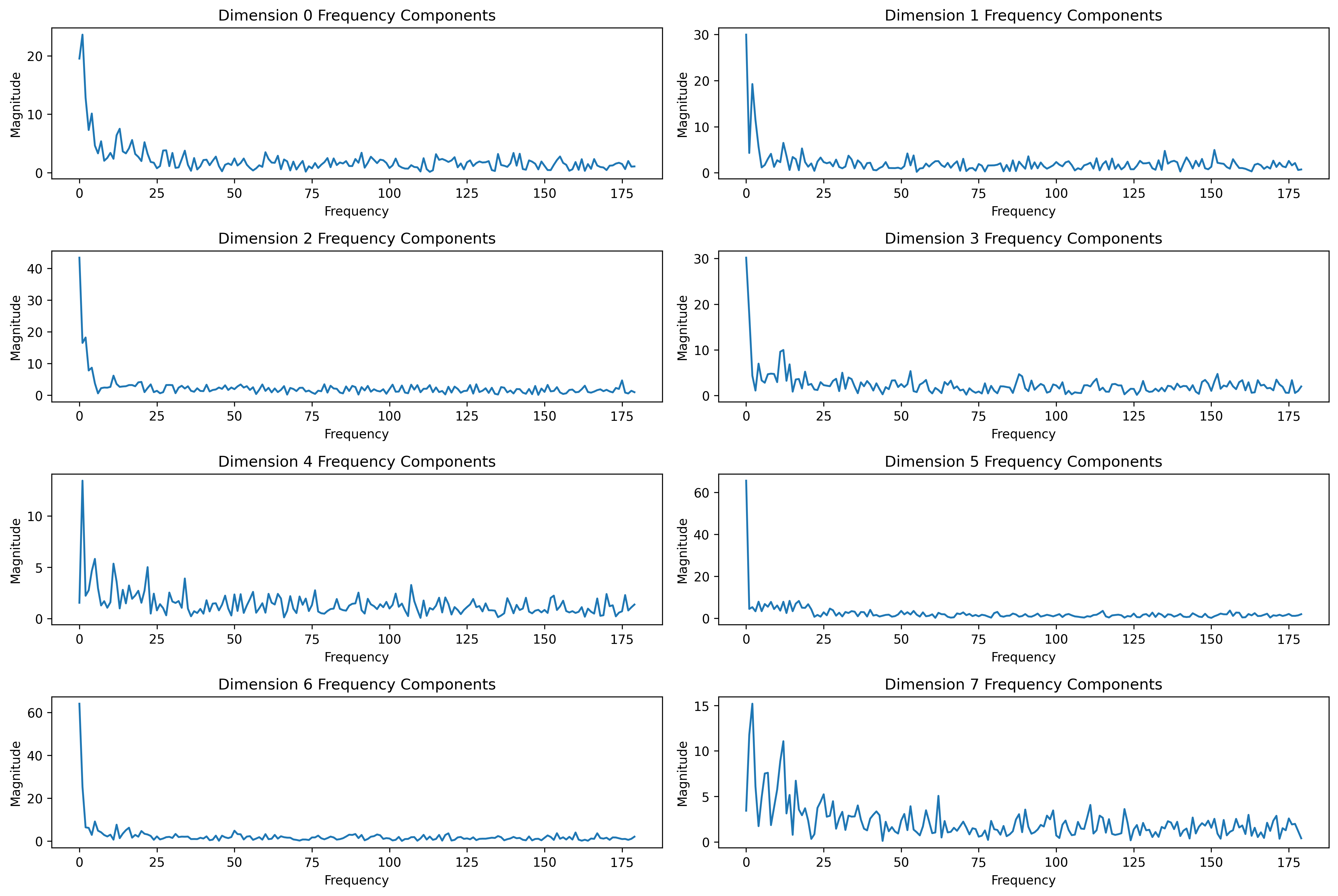}
\caption{Frequency spectrum analysis: (Left) Sinusoidal embeddings distribute energy across all frequencies. (Right) Learned embeddings concentrate in low frequencies ($<$ 0.1 cycles/day), suggesting the model prioritizes broad temporal context over precise positioning.}
\label{fig:fft_analysis}
\end{figure}

\subsection{Key Findings}

Learned embeddings concentrate $>$80\% of their energy in frequencies below 0.1 cycles/day (periods $>$10 days). High-frequency components ($>$0.5 cycles/day) are nearly absent.This frequency distribution suggests that for user behavior prediction, distinguishing "last week" from "last month" is more important than distinguishing "day 42" from "day 45". The model learns to encode coarse temporal granularity rather than precise day-level positioning. Fixed sinusoidal embeddings distribute energy uniformly across all frequencies, potentially encoding unnecessary high-frequency information. The learned embeddings adaptively focus on task-relevant temporal scales.

\subsection{Comparison of Positional Encoding Strategies}

We systematically evaluated five positional encoding strategies on the user behavior prediction task:

\begin{table}[h]
\centering
\caption{Performance comparison of positional encoding strategies}
\label{tab:positional}
\begin{tabular}{lcc}
\toprule
Strategy & Validation Loss & Description \\
\midrule
None & 0.3412 & No positional information \\
Sinusoidal & 0.3389 & Fixed trigonometric functions \\
RoPE & 0.3365 & Rotary position embeddings \\
TAPE & 0.3358 & Time-aware position embeddings \\
\textbf{Learned} & \textbf{0.3347} & Trainable embeddings (used in LUMOS) \\
\bottomrule
\end{tabular}
\end{table}

\section{Embedding Extraction and Analysis}
\label{sec:embeddings}

\textbf{Embedding Quality}: We extracted user embeddings from the trained multi-task model according to the method described in Section \ref{sec:extracting-user-embeddings} and evaluated their predictive performance on downstream tasks to assess generalizability.

\subsection{Extracting User Embeddings}
\label{sec:extracting-user-embeddings}

Beyond predictions, LUMOS generates valuable user embeddings for downstream tasks. We extract representations at multiple levels:

\begin{itemize}
    \item \textbf{Static embedding}: $\mathbf{h}^{static}$ for cold-start scenarios with minimal user information
    \item \textbf{Dynamic User embedding}: Aggregated encoder output capturing current temporal user state
    \item \textbf{Event context embedding}: Learned representation of event context
\end{itemize}

\subsubsection{Dynamic User Embeddings}

A key innovation in LUMOS is the generation of dynamic user embeddings that are refreshed daily. These embeddings capture the evolving state of each user based on their most recent behavioral patterns:

\begin{equation}
\mathbf{e}^{user}_t = \text{Aggregate}(\mathbf{H}^{enc}_t)
\end{equation}

where $\mathbf{H}^{enc}_t$ is the encoder output at time $t$ incorporating all historical information up to that point.

For aggregation, we primarily use exponentially weighted averaging:
\begin{equation}
\mathbf{e}^{user}_t = \sum_{i=1}^{T_{hist}} \alpha^{T_{hist}-i} \mathbf{H}^{enc}_{t,i}
\end{equation}

where $\alpha \in (0,1)$ is a decay factor that emphasizes recent history. This approach ensures that the embedding reflects the user's current state while retaining important historical patterns.

These dynamic embeddings serve multiple purposes, including identifying cohorts with similar behavioral patterns through user similarity detection, flagging sudden changes in user behavior via anomaly detection, tailoring content and offers based on the current user state through personalization, and providing rich inputs for downstream ML models via feature extraction.

\subsection{User Embedding Analysis}

\subsubsection{Embedding Quality on Downstream Tasks}

We evaluate learned user embeddings on five downstream tasks using simple logistic regression:

\begin{table}[h]
\centering
\caption{User embedding performance on downstream tasks (ROC-AUC)}
\label{tab:embeddings}
\begin{tabular}{lccccc}
\toprule
Embedding Source & Task 1 & Task 2 & Task 3 & Task 4 & Task 5 \\
\midrule
Single-task (Task 1 only) & 0.906 & 0.654 & 0.411 & 0.657 & 0.974 \\
Single-task (Task 2 only) & 0.879 & 0.672 & 0.425 & 0.641 & 0.961 \\
\textbf{Multi-task LUMOS} & \textbf{0.977} & \textbf{0.722} & \textbf{0.442} & \textbf{0.708} & \textbf{0.975} \\
\midrule
Relative Improvement & +7.8\% & +7.4\% & +4.0\% & +7.8\% & +0.1\% \\
\bottomrule
\end{tabular}
\end{table}

Multi-task embeddings dramatically outperform single-task specialists, even on the tasks the specialists were trained for. This suggests that joint training forces the model to learn more generalizable representations that capture fundamental aspects of user behavior rather than task-specific patterns. We found that exponentially weighted aggregation with high decay factor performs best, confirming that recent behavior is most predictive while historical context provides important background.

\subsection{Event Context Embedding Analysis}

The learned event context embeddings ($d_{supply\_embed}=64$) capture meaningful event context. Each day's event context features (match counts, tournament indicators) are embedded into a continuous vector space where semantically similar days cluster together.

\begin{figure}[htbp]
\centering
\includegraphics[width=0.7\columnwidth]{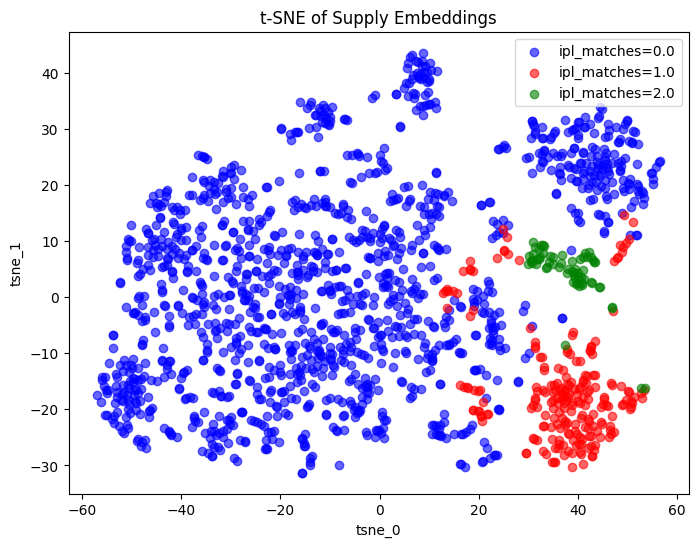}
\caption{t-SNE projection of daily event context embeddings reveals distinct clusters for days with different IPL match counts: (blue) no IPL matches, (orange) single IPL match, (green) IPL double-headers. The learned space captures event context similarity without explicit supervision.}
\label{fig:supply_tsne}
\end{figure}

\textbf{Use Cases for Event Context Embeddings:}
\begin{itemize}
    \item \textbf{Event Context Similarity Analysis}: Find historical days with similar event calendars
    \item \textbf{What-If Scenarios}: Embed hypothetical future event context to simulate user responses
    \item \textbf{Event Impact Assessment}: Measure distance between normal days and tournament days
    \item \textbf{Communications Targeting}: Identify users most responsive to specific event context patterns
\end{itemize}

\subsection{Ablation Study: Impact of Event Context Data}

We investigate the importance of event context by systematically removing past and future event context information:

\begin{table}[h]
\centering
\caption{Event context ablation results on churn prediction}
\label{tab:supply_ablation}
\begin{tabular}{lccc}
\toprule
Configuration & Validation Loss & ROC-AUC & $\Delta$ Loss \\
\midrule
Full Model (Past + Future Event Context) & \textbf{0.3331} & \textbf{0.9300} & - \\
No Future Event Context & 0.3350 & 0.9294 & +0.57\% \\
No Past Event Context & 0.3349 & 0.9297 & +0.54\% \\
No Event Context (Past or Future) & 0.3345 & 0.9298 & +0.42\% \\
\bottomrule
\end{tabular}
\end{table}

Surprisingly, removing either past or future event context individually hurts performance more than removing both. This suggests that past and future event context provide complementary information—the model learns to associate historical user responses to past events with likely responses to similar future events. When only one is available, the model cannot make these connections effectively.

% Implementation appendices
\section{Production Inference Infrastructure}
\label{sec:inference}

\subsection{Batch Inference Architecture}

LUMOS operates in daily batch inference mode, scoring all active users ($\sim$5M/day) each morning. The inference pipeline consists of five stages:

\begin{figure}[htbp]
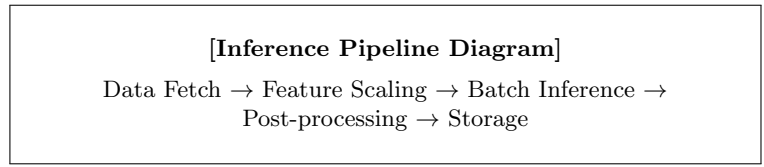

\centering
\fbox{\parbox{0.8\columnwidth}{\centering
\vspace{1em}
\textbf{[Inference Pipeline Diagram]}\\[0.5em]
Data Fetch $\rightarrow$ Feature Scaling $\rightarrow$ Batch Inference $\rightarrow$ Post-processing $\rightarrow$ Storage\\
\vspace{1em}
}}
\caption{Production inference pipeline: Data fetch $\rightarrow$ Feature scaling $\rightarrow$ Batch inference $\rightarrow$ Post-processing $\rightarrow$ Storage.}
\label{fig:inference_arch}
\end{figure}

\appendix

\section{Implementation Details}

% Infrastructure section to be rewritten

\end{document}